\documentclass{article}

    \PassOptionsToPackage{numbers, compress}{natbib}
    \usepackage[preprint]{neurips_2023}

\usepackage[utf8]{inputenc} % allow utf-8 input
\usepackage[T1]{fontenc}    % use 8-bit T1 fonts
\usepackage[colorlinks,
linkcolor=red,
anchorcolor=cyan,
citecolor=cyan,
]{hyperref}       % hyperlinks
\usepackage{url}            % simple URL typesetting
\usepackage{booktabs}       % professional-quality tables
\usepackage{amsfonts}       % blackboard math symbols
\usepackage{nicefrac}       % compact symbols for 1/2, etc.
\usepackage{microtype}      % microtypography
\usepackage{xcolor}         % colors
\usepackage{multirow}
\usepackage{amsmath}
\usepackage{bm}
\usepackage{amssymb}
\usepackage[bottom]{footmisc}
\usepackage{amsthm}
\usepackage{amsmath}

\newtheorem{theorem}{Theorem}

\usepackage{graphicx}
\usepackage{float}
\usepackage{caption}
\usepackage{makecell}
\usepackage{wrapfig}

\definecolor{ggreen}{RGB}{15,157,88}

\title{Understanding the Effect of Data Augmentation on Knowledge Distillation}

\author{%
  Ziqi Wang$^1$, Chi Han$^1$, Wenxuan Bao$^1$, Heng Ji$^1$ \\
  $^1$ Department of Computer Science, University of Illinois Urbana-Champaign \\
  \texttt{\{ziqiw9, chihan3, wbao4, hengji\}@illinois.edu} \\
}

\begin{document}

\maketitle

\begin{abstract}
    Knowledge distillation (KD) requires sufficient data to transfer knowledge from large-scale teacher models to small-scale student models. Therefore, data augmentation has been widely used to mitigate the shortage of data under specific scenarios. Classic data augmentation techniques, such as synonym replacement and k-nearest-neighbors, are initially designed for fine-tuning. To avoid severe semantic shifts and preserve task-specific labels, those methods prefer to change only a small proportion of tokens (e.g., changing $10\%$ tokens is generally the best option for fine-tuning). However, such data augmentation methods are sub-optimal for knowledge distillation since the teacher model could provide label distributions and is more tolerant to semantic shifts. We first observe that KD prefers as much data as possible, which is different from fine-tuning that too much data will not gain more performance. Since changing more tokens leads to more semantic shifts, we use the proportion of changed tokens to reflect semantic shift degrees. Then we find that KD prefers augmented data with a larger semantic shift degree (e.g., changing $30\%$ tokens is generally the best option for KD) than fine-tuning (changing $10\%$ tokens). Besides, our findings show that smaller datasets prefer larger degrees until the out-of-distribution problem occurs (e.g., datasets with less than 10k inputs may prefer the $50\%$ degree, and datasets with more than 100k inputs may prefer the $10\%$ degree). Our work sheds light on the preference difference in data augmentation between fine-tuning and knowledge distillation and encourages the community to explore KD-specific data augmentation methods.
\end{abstract}

\section{Introduction}

\begin{wrapfigure}{r}{0.45\textwidth}
    \centering
    \resizebox{0.45\textwidth}{!}{
    \includegraphics{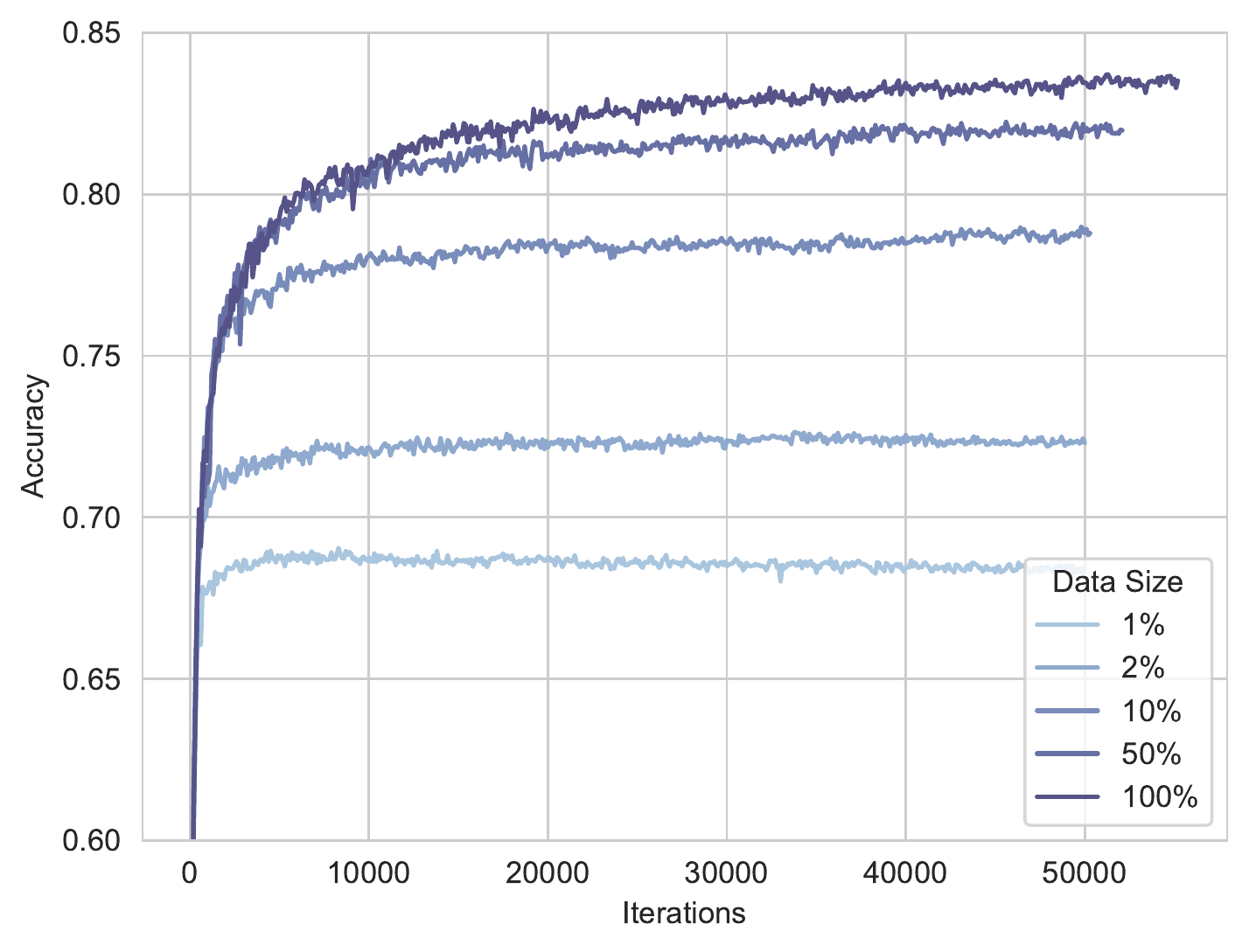}
    }
    
    \caption{KD from a teacher model BERT (Large) to a student model DistilBERT (Base) on MNLI dataset. The teacher model is fine-tuned on the whole dataset. Results show that more distillation data will benefit distillation performance.}
    
    \label{fig:kd_needs_data}
\end{wrapfigure}

Knowledge distillation (KD) \citep{hinton2015distilling} is a widely used framework to transfer knowledge between models. KD benefits real-world applications by distilling knowledge from large teacher models to small student models, such as deploying language models on edge devices \citep{sun2020mobilebert}. However, KD usually requires a large amount of data (see Figure \ref{fig:kd_needs_data} for an example), which cannot be met in many real-world scenarios such as biomedical domain \cite{herrero2013ddi, bravo2015extraction, van2012eu, krallinger2017overview}.

Recently various data augmentation methods have been proposed, including synonym replacement (SR) \citep{kolomiyets-etal-2011-model}, k-Nearest-Neighbors (kNN) \citep{wang2015s}, and easy data augmentation (EDA for short), which combines four lexical level augmentation methods: synonym replacement, random swap, random insertion, and random deletion \citep{wei-zou-2019-eda}. These approaches have been proven useful to enlarge the data and help knowledge distillation \citep{feng2021learning, wang2023augmentation}. These methods introduce little computational overhead, thus, are suitable for augmenting a large amount of data in a short time. Autoregressive
models (e.g., T5 \cite{raffel2020exploring}, GPT-3 \cite{brown2020language}, ChatGPT \cite{openai2022chatgpt}) can generate high-quality synthesized data \citep{yoo-etal-2021-gpt3mix-leveraging, zhou-etal-2022-flipda} but are much more computationally expensive than classic data augmentation methods (e.g., synonym replacement), which limits the size of synthesized data. Consequently, synthesized data generated by autoregressive models are preferred in the few-shot setting \citep{yoo-etal-2021-gpt3mix-leveraging, zhou-etal-2022-flipda} rather than the full-data setting.

Although these popular data augmentation methods can be used directly in knowledge distillation, they are initially designed for vanilla fine-tuning or few-shot settings. For example, SR and EDA tend to only change a small proportion of tokens in the text to avoid severe semantic shifts and preserve task-specific labels, \cite{zhou-etal-2022-flipda} uses T5 \citep{raffel2020exploring} to change a few tokens of the original input to generate a new input that flips the label. In knowledge distillation, we are not required to know the labels of augmented data since the teacher model can provide the label distribution. Therefore, these data augmentation methods may be sub-optimal for knowledge distillation. In this paper, we hope to shed light on the following question:

\texttt{What data augmentation is the best fit for knowledge distillation?}

This question contains two sub-questions: (1) \textbf{which data augmentation paradigm does KD prefer?} More specifically, should we use higher-quality but lower-quantity synthesized data \citep{yoo-etal-2021-gpt3mix-leveraging} generated by autoregressive models or lower-quality but higher-quantity augmented data generated by classic data augmentation methods \citep{kolomiyets-etal-2011-model, wang2015s, wei-zou-2019-eda}? We call the first type "synthesized data" and the second type "augmented data" to simplify the notion. (2) \textbf{how should we make the data augmentation paradigm perform well in KD?} The first sub-question is about selecting a paradigm; the second is about adapting a paradigm initially designed for non-KD settings to KD. 

For the first sub-question, our experiments (Section \ref{sec:sub1}) show that quantity is more important than quality from the perspective of KD. This observation encourages us to use augmented data that is cheap and can be deployed in an online manner (so that we can obtain a large amount of augmented data without the concern of storage). This property is different from fine-tuning, which has been shown that more data will not result in more performance gain \citep{wei-zou-2019-eda}. We answer the second sub-question (Section \ref{sec:sub2}) from the observation that knowledge distillation is more tolerant to semantic shifts and can provide the label probability distribution for arbitrary inputs. Previous data augmentation methods find that changing a small proportion of tokens gives the best empirical results (e.g., 10\% tokens for fine-tuning \citep{wei-zou-2019-eda}), which may not be optimal for KD. Intuitively, changing more tokens can cause more semantic shifts, produce more diverse data, thus fully utilize teacher models. Since the proportion of changed tokens reflects the degree of semantic shift, we use "semantic shift degree" or "degree" to denote the proportion of changed tokens to simplify the notion. We theoretically (Section \ref{sec:sub2}) and empirically (Section \ref{sec:exp}) find that a smaller data size prefers a more significant semantic shift degree and show that $30\%$ degree (i.e., $30\%$ tokens are changed) produces the best KD results in most cases. At last, we show how our findings benefit the real-world biomedical application (Section \ref{sec:app}).

% \vspace{-2mm}

Our findings can be summarized as follows:
\begin{itemize}
    \item In KD, augmented data is preferable to synthesized data due to the low cost. The quantity of augmented data is important to ensure well-distilled student models. Therefore, KD encourages cheap augmented data and online augmentation.
    \item Augmented data should have a larger semantic shift degree to fully utilize teacher models and achieve a better KD performance. Generally speaking,  the $30\%$ degree is a "sweet spot" in KD (our findings), and 10\% degree is a "sweet spot" in fine-tuning \cite{wei-zou-2019-eda}.
    \item Our experiments and theoretical analysis show that smaller datasets prefer a larger degree.
    \item Though KD prefers a larger semantic shift degree, an extremely large degree is not a good option since it will cause an out-of-distribution (OOD) problem and thus hurt the KD performance.
\end{itemize}

Our work sheds light on the difference between KD and fine-tuning from the perspective of data augmentation and gives guidance on how to use data augmentation for KD. By adapting current data augmentation methods to the KD setting, we show the potential power of data augmentation in knowledge distillation. Therefore, we encourage the community to explore more KD-specific data augmentation methods.

\section{Background}
Knowledge distillation aims to distill knowledge between models. Formally speaking, given a dataset $\mathcal{D} = \{(x,y)\}$, a fine-tuned teacher model $f_t: \mathcal{X} \rightarrow \mathcal{Y}$, and student model to be trained $f_g: \mathcal{X} \rightarrow \mathcal{Y}$, knowledge distillation train the student model with a loss

$$\mathcal{L}_{KD} = \sum_\mathcal{D} \alpha\ \mathcal{L}_{ft}(f_g(x),y) + (1-\alpha)\ \text{Distance}(f_g(x),f_t(x))$$

where $\mathcal{L}_{ft}$ denotes the fine-tuning loss (e.g., cross-entropy) and $\text{Distance}$ measures the distance between $f_g(x)$ and $f_t(x)$ (e.g., KL divergence and mean-square-error). $\alpha \in [0,1]$ is the weight between two losses. Empirically, smaller $\alpha$ leads to better KD performance. Augmented data (or synthesized data) $\mathcal{D'} = \{(x',y')\}$ could also be used in KD

\begin{align*}
\mathcal{L'}_{KD} &= \sum_\mathcal{D'} \beta\ \mathcal{L}_{ft}(f_g(x'),y') + (1-\beta)\ \text{Distance}(f_g(x'),f_t(x')) \\
\mathcal{L} &= \gamma\ \mathcal{L}_{KD} + (1-\gamma)\ \mathcal{L'}_{KD}    
\end{align*}

$\beta, \gamma \in [0,1]$ control weights among losses. Classic data augmentation methods, such as synonym replacement, only generate $x'$ and assume $y'=y$. Autoregressive models such as GPT-3 can generate $x'$ as well as $y'$.

\section{Which Data Augmentation Paradigm Does KD Prefer?}
\label{sec:sub1}

Data augmentation methods could be divided into two paradigms based on their quality and efficiency. (1) Augmented data (i.e., lower-quality but higher-quantity): SR, kNN, and EDA are all in this paradigm. The main feature of these types of data is that they are very cheap to be produced. In practice, we can easily obtain a large amount of such data. (2) synthesized data (i.e., higher-quality but lower-quantity): Large language models could help generate high-quality synthesized data but with a high computational expense.

We conduct an experiment to compare their performance in KD. To get stable results and diminish the effect of variance, we choose MNLI (392.7k) \citep{N18-1101} as our benchmark. We fine-tune a BERT (Large) (336M) \citep{devlin-etal-2019-bert} on the full MNLI dataset as the teacher model. DistilBERT (Base) (66M) \citep{sanh2019distilbert} is selected as the student model. To highlight the effect of data augmentation, we only use $1\%$ of the MNLI training data (i.e., $\sim$ 3900 input-output pairs) and evaluate the KD on the MNLI-M validation set. All experiments in the paper are running on 1x V100 16GB unless stated otherwise.

We choose kNN as a representative of augmented data. kNN generates augmented data by randomly replacing $r\%$ tokens to one of its top-k neighbors in the embedding space. We use the default hyperparameters used in previous fine-tuning work \citep{zhou-etal-2022-flipda}, i.e., we set $k=15$ and $r=10$. Since kNN is cheap, there are two ways of augmentation. Offline augmentation will first augment a fixed amount of data (4x and 8x KD data in our experiments) and mix augmented data with KD data. Online augmentation will augment the same amount of data as the batch size whenever we sample a batch from KD data. Therefore, online augmentation could augment more data with more distillation steps. Besides, we do not need to store augmented data with online augmentation. In our experiments, we set the batch size to 64 and distillation steps to 50,000, therefore online augmenting total $64 * 50,000 / 3900 = 820.5$x KD data.

We use back translation \citep{edunov2018understanding} to generate synthesized data. To ensure the data is of high quality, we choose NLLB 1.3B \citep{costa2022no}, one of the largest machine translation models that support translation between 200 languages. To ensure the diversity of data, we use eight target languages
\footnote{ace\_Arab, fra\_Latn, hin\_Deva, jpn\_Jpan, rus\_Cyrl, shn\_Mymr, tha\_Thai, zho\_Hans}
in total. Each target language generates the same amount of synthesized data as KD data. Therefore, the synthesized data is 8x times the KD data. We do not choose ChatGPT or GPT-3 for due to the slow response of APIs, which will take an unbearable time to generate enough data. However, it is possible to estimate the upper-bound performance of ChatGPT by assuming the synthesized data is the ground-truth data in the original dataset. The time cost of ChatGPT can be estimated by the API call time cost.

Figure \ref{fig:syn_vs_aug} and Table \ref{tab:cost} show the KD performance with different data augmentation methods and the time costs. We can conclude that (1) More data improves KD performance more. The kNN and back translation help KD perform better with more data, which is different from fine-tuning that too much data is not helpful \citep{wei-zou-2019-eda}. Figure \ref{fig:syn_vs_aug_ft} shows the fine-tuning results under the same setting as KD. We observe that more data will not lead to performance gain (BT 4x vs. BT 8x, kNN 4x vs. kNN 8x vs. kNN online) in fine-tuning, but high-quality data helps performance improvement, which differs from KD. Moreover, we can see that kNN and BT do not help improve performance and even have a performance drop with increased iterations in fine-tuning. This may be because the assumption that $y'=y$ may not hold and introduces noises. In the smaller dataset ($1\%$ MNLI training data only has $\sim 3900$ inputs), the noise could affect the performance more significantly, causing a performance drop. (2) KD prefers augmented data more than synthesized data. Synthesized data performs better than augmented data when they are of the same amount (4x, 8x); augmented data outperforms synthesized data when their computation costs are the same. The online kNN augmentation takes only 103 mins (and less GPU usage) but outperforms back translation 8x (112 mins) and ChatGPT 1x (260 mins). Besides, we could also find that online augmentation has not converged yet (unlike offline augmentation, which converges quickly and stays steady), meaning the online augmentation's performance could still be improved with increased iterations. \textcolor{ggreen}{\textbf{Based on the above observations, we suggest using augmented data in an online manner for KD.}}

\begin{table}
\centering
\caption{Time cost of different data augmentation methods. ChatGPT time cost is an estimation. In our experiments, we found the speed of API calls is 1300 tokens/40s. We assume each instance has 128 tokens.}
\vspace{1mm}
    \resizebox{0.6\textwidth}{!}{
    \begin{tabular}{c|ccc}
       \textbf{Methods}  & \textbf{kNN (k=15)} & \textbf{Back Translation} & \textbf{ChatGPT} \\
       \midrule
       \textbf{Hardware} & 1x V100 16GB & 1x V100 16GB & API \\
       \textbf{\# Augmented Data} & 4x/8x/820.5x(online) & 4x/8x  & 1x\\
       \textbf{Time Cost (min)} & 0.5/1/103 & 56/112 & 260$^*$\\
    \end{tabular}
    }
    \label{tab:cost}
\end{table}

\begin{minipage}[t]{\textwidth}
\begin{minipage}[t]{0.48\textwidth}

     \includegraphics[width=\textwidth]{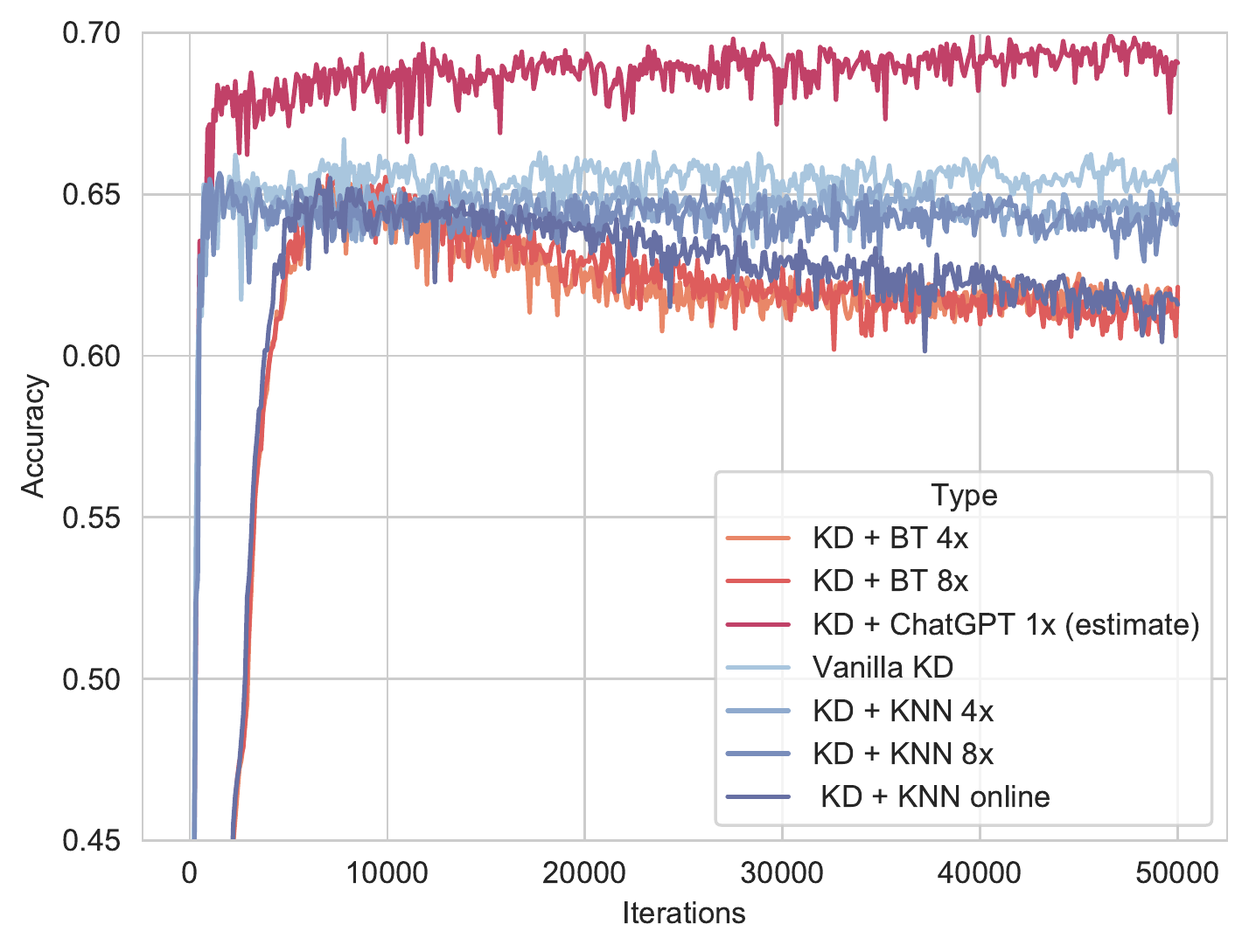}
    \captionof{figure}{Fine-tuning performance with different data augmentation methods. BT denotes back translation, and Nx indicates that the augmented data is N times the number of original data.
    }
    \label{fig:syn_vs_aug_ft}

\end{minipage}
\hfill
\begin{minipage}[t]{0.48\textwidth}
    \includegraphics[width=\textwidth]{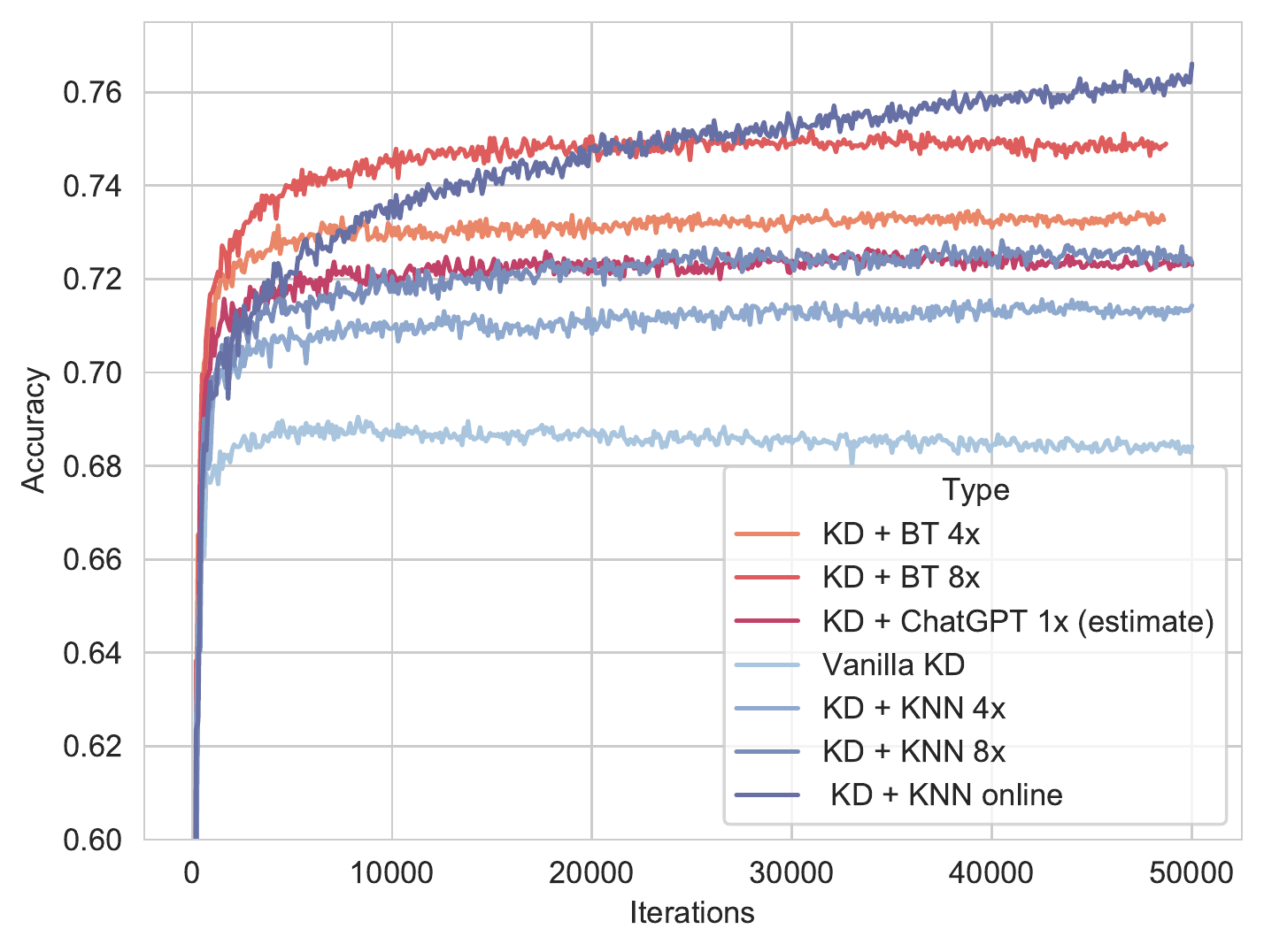}
    \captionof{figure}{KD performance with different data augmentation methods. BT denotes back translation, and Nx indicates that the augmented data is N times the number of original data.
    }
    \label{fig:syn_vs_aug}

\end{minipage}
\end{minipage}
\section{Adapt Augmented Data to KD}
\label{sec:sub2}
\subsection{Intuition}
Although augmented data fits KD better than synthesized data, they may still be sub-optimal. Fine-tuning requires manual labels. Therefore, augmented data cannot be too far away from the original data (e.g., $10\%$ semantic shift degree is generally the best option in fine-tuning \citep{wei-zou-2019-eda}) to preserve task-specific labels. In contrast, label distributions can be provided by teachers in knowledge distillation. Intuitively, a higher semantic shift degree will produce more diverse data and fully utilize teachers' abilities. It is also worth noting that an extremely large semantic shift degree is not an ideal solution because of out-of-distribution (OOD) problems. An extreme case is that we could randomly sample several tokens from a vocabulary and formulate them into natural language sentences to do KD, leading to lousy distillation performance \cite{wang2023augmentation}.

To sum up, we could expect that an appropriate large semantic shift degree in KD will lead to better distillation performance, whereas a smaller or extremely large degree will lead to worse performance (sub-optimal and OOD). This intuition could be understood theoretically.

\subsection{Theory Understanding}

\paragraph{Setting}
In this work, we study a formal framework for measuring the effect of data augmentation in KD. We assume a ground-truth data distribution $\mathcal{D}=\{(x, y)\}$ with $x\in \mathbb{R}^d \sim \mathcal{N}(0, I)$ where $d\geq 2$, and $y \sim \mathcal{B}(f(x))$. We also assume a Lipschitz bound of B on $P(x)$. Training data $ D_{train}$ and testing data $\mathcal{D}_{test}$ are $n$ and $m$ i.i.d. sample from $\mathcal{D}$, respectively. We also assume access to a teacher model $g(x)$, so that $\forall x, |g(x)-f(x)|<\epsilon_t$. The teacher model does not have information on input density $p(x)$.
Assume for each training sample $(x,y)$, we augment its information to span a local Gaussian distribution $\mathcal{N}(x, \tau I)$. We then conduct an augmentation to the dataset: we i.i.d. sample $m$ datum points from the following distribution:

\begin{align*}
    \mathcal{P}_{aug}(x') &= \frac{1}{n}
    \sqrt{\frac{1}{(2\pi)^d\tau}}
    \sum_{x\in D_{train}}
    e^{-\frac{\|x'-x_i\|_2^2}{2\tau}},\ \ \ \ y'=g(x'),
\end{align*}

and get the augmented dataset $\mathcal{D}_{aug}$.
Then we formalize the above intuition and propose the following theorem:

\begin{theorem}
\label{thm:teacher_augmentation}

Let $x_u=2d\ln(2\pi d) + 12\ln(2) - 4\ln(\epsilon)$, as long as

\begin{align*}
    \tau <& \frac{2^{d-3}\epsilon}{x_u^{d+2}},\ \ \ \ n > \frac{16}{\epsilon^2} \left((d^2+5d+6)\ln2 + (d^2+d)\ln x_u + d\ln d - d\ln\epsilon - d\ln\delta-\frac{d}{2}\ln\tau \right)
\end{align*}

we have with $1-\delta$ probability,
for any $h\in\mathcal{H}, \epsilon_T(h) - \epsilon_S(h) < \epsilon + \lambda$ where $\lambda=\epsilon_T(h^\star)+\epsilon_T(h^\star)$.
\end{theorem}

The proof can be found in Appendix \ref{appendix:thm}. Some takeaways for this theorem:
\begin{itemize}
    \item $\tau$ can be regarded as the semantic shift degree. Higher $\tau$ means a larger degree (i.e., more tokens to be changed).
    \item The best choice of $\tau$ is given by $n$ (data sizes). Smaller dataset prefers higher $\tau$.
    \item Lower $\tau$ loses the error bound provided by the theorem; higher $\tau$ introduces the OOD problem and breaks the theorem by failing to meet the condition $\forall x, |g(x)-f(x)|<\epsilon_t$, leading to higher errors.
\end{itemize}

In practice, the theorem encourages us to augment data with a higher semantic shift degree in KD (but not extremely large), and the less data we have, the higher degree we should use.
\begin{table}
    \centering
    \caption{Results of KD performance (Accuracy $\%$) with different KD data sizes and semantic shift degrees of kNN.}
        \vspace{1mm}
    \resizebox{\textwidth}{!}{
    \begin{tabular}{c|ccccccccc}
    \toprule
    \multirow{2}{*}{\textbf{KD Data Sizes}} &\multicolumn{9}{|c}{\textbf{Semantic Shift Degree}} \\
     & \textbf{0\% (Vanilla KD)} & \textbf{10\%} & \textbf{20\%} & \textbf{30\%} & \textbf{40\%} & \textbf{50\%} & \textbf{60\%} & \textbf{70\%} & \textbf{80\%} \\
     \midrule
      \multicolumn{10}{c}{\textbf{MNLI (392.7k) (Evaluated on the MNLI-M validation set)}} \\
    \midrule
     \textbf{1\%} & $68.89$ & $76.39$	& $77.31$& 	$77.17$	& $\textcolor{ggreen}{\bm{78.06}}$	&$77.47$	&$77.02$&	$76.23$&	$75.31$\\
     \textbf{2\%} & $72.58 $ & $ 77.81	$ & $78.87$ & $	\textcolor{ggreen}{\bm{79.24}}	$ & $79.00$ & $	78.84	$ & $78.22$ & $	77.57	$ & $76.88$\\
     \textbf{10\%} & $78.88$ & $ 80.69	$ & $81.35	$ & $\textcolor{ggreen}{\bm{81.53}}	$ & $81.46	$ & $81.2$ & $	80.96$ & $	80.55	$ & $80.21$\\
     \textbf{100\%} & $83.65$ & $ \textcolor{ggreen}{\bm{84.14}}$ & $	84.05	$ & $84.08	$ & $83.96	$ & $83.85	$ & $83.63	$ & $83.8	$ & $83.45$\\
    \midrule
    \multicolumn{10}{c}{\textbf{QNLI (104.7k)}} \\
    \midrule
    \textbf{3\%} & $77.86 $ & $81.47	$ & $83.44$ & $	84.32$ & $	84.67$ & $	\textcolor{ggreen}{\bm{85.58}}$ & $	85.01$ & $	84.24$ & $	83.8$\\
    \textbf{6\%} & $81.2$ & $ 84.33$ & $	85.87$ & $	86.04$ & $	86.67$ & $	\textcolor{ggreen}{\bm{87.19}}$ & $	86.19$ & $	85.67	$ & $84.49$\\
    \textbf{20\%} & $82.31$ & $ 85.51$ & $	87.45$ & $	87.69$ & $	\textcolor{ggreen}{\bm{88.12}}	$ & $87.68$ & $	87.79$ & $	86.87$ & $	86.31$\\
    \textbf{100\%} &$88.07$ & $ 89.23	$ & $89.82$ & $	90.22	$ & $\textcolor{ggreen}{\bm{90.39}}	$ & $90.07	$ & $90.33	$ & $90.28	$ & $89.84$ \\
    \bottomrule
    \end{tabular}
    }
    \label{tab:exp1}
\end{table}

\section{Experiments}
\label{sec:exp}
We design experiments to (1) show the performance of KD with different semantic shift degrees, (2) show how data sizes affect the choice of semantic shift degrees, (3) and give an empirical suggestion for choosing the degree.

\begin{minipage}{0.43\textwidth}
% \begin{table}[!htbp]
    \centering
    \captionof{table}{The student model's fine-tuning performance with different data sizes and semantic shift degrees of kNN.}
    % \vspace{1mm}
    \resizebox{\textwidth}{!}{
    \begin{tabular}{c|ccccc}
    \toprule
    \multirow{2}{*}{\textbf{FT Data Sizes}} &\multicolumn{5}{|c}{\textbf{Semantic Shift Degree}} \\
     & \textbf{0\% (Vanilla FT)} & \textbf{10\%} & \textbf{30\%}  & \textbf{50\%}  & \textbf{70\%}  \\
     \midrule
      \multicolumn{6}{c}{\textbf{MNLI (392.7k) (Evaluated on the MNLI-M validation set)}} \\
    \midrule
     \textbf{1\%} & $66.69 $ & $\textcolor{ggreen}{\bm{65.86}}$ & $	65.78	$ & $	65.84$ & $	65.79$\\
     \textbf{2\%} & $69.91$ & $ \textcolor{ggreen}{\bm{68.91}}$ & $	68.61	$ & $	68.47$ & $	68.67$\\
     \textbf{10\%} & $75.22$ & $ \textcolor{ggreen}{\bm{75.20}}$ & $	75.19	$ & $	75.19$ & $	75.12$\\
     \textbf{100\%} &$81.92$ & $ \textcolor{ggreen}{\bm{82.27}}$ & $	82.08		$ & $82.09	$ & $81.74$\\
    \bottomrule
    \end{tabular}
    }
    \label{tab:exp_ft}
% \end{table}
\end{minipage}
\hfill
\begin{minipage}{0.55\textwidth}
% \begin{table}[!htbp]
    \centering
    \captionof{table}{More dataset results of KD with different semantic shift degrees.}
    % \vspace{1mm}
    \resizebox{\textwidth}{!}{
    \begin{tabular}{c|ccccc}
    \toprule
    \multirow{2}{*}{\textbf{Dataset}} &\multicolumn{5}{|c}{\textbf{Semantic Shift Degree}} \\
     & \textbf{0\% (Vanilla KD)} & \textbf{10\%} & \textbf{30\%}  & \textbf{50\%}  & \textbf{70\%}  \\
    \midrule
     \textbf{CoLA (8.5k) (Matthew)} & $53.95 $ & $57.97$ & $	59.05	$ & $	\textcolor{ggreen}{\bm{59.78}}$ & $	59.13$\\
     \textbf{MRPC (3.7k) (Acc)} & $85.13 $ & $85.52$ & $	86.61	$ & $	86.44$ & $	\textcolor{ggreen}{\bm{87.03}}$\\
     \textbf{RTE (2.5k) (Acc)} & $67.2$ & $ 67.16$ & $	\textcolor{ggreen}{\bm{67.54}}$ & $		67.26$ & $	65.73$\\
     \textbf{QQP (363.8k) (Acc)} &$ 90.55$ & $\textcolor{ggreen}{\bm{90.8}}	$ & $90.75$ & $		90.59	$ & $90.69$\\
    \bottomrule
    \end{tabular}
    }
    \label{tab:exp1.2}
% \end{table}
\end{minipage}

Similar to Section \ref{sec:sub1}, we use kNN as our data augmentation baseline ($k=15$, online augmentation), BERT (Large) fine-tuned on the whole dataset as the teacher, and DistilBERT (Base) as the student. We use the same random seed for all experiments. Mean-square-error is used as the $\text{Distance}$ function. We set $\alpha=0.1, \beta=0.0, \gamma=0.5$ in our experiments. We do not use $y'$ ($\beta=0.0$) since kNN assumes $y'=y$, which may be false and introduce noises. We run experiments with 0.01 and 0.001 learning rates and report the best result. The Batch size is 64 and the total distillation steps is 50,000.

\subsection{KD prefers a larger semantic shift degree}
\label{sec:kdprefer}
To show how the semantic shift degrees of kNN affect KD performance, we set the degree from $10\%$ to $80\%$ and data size from $1\%$ from $100\%$. Results are shown in Table. \ref{tab:exp1}. We can observe that KD prefers larger degrees. $30\%$ $\sim$ $50\%$ degrees are generally the best fit for KD, whereas the $10\%$ degree is always the best fit for fine-tuning when using data augmentation (Table \ref{tab:exp_ft}). Moreover, we can see that larger degrees ($30\%$ $\sim$ $50\%$ ) significantly improve the KD performance compared with the smaller degrees ($10\%$). On the MNLI dataset, the performance gain differs by $2\%$ when the data size is small ($1\%\sim2\%$). On the QNLI dataset, the performance gain differs from $3\%$ to $4\%$, showing the consistent benefit of larger degrees. Another observation is that the model performance drops dramatically when degrees become extremely large, which is expected since the existence of the OOD problem. We can also conclude that KD prefers larger degrees when the dataset is smaller, which aligns with our theory. In the MNLI dataset, $40\%$ degree is the best fit for the $1\%$ dataset, $30\%$ degree is the best fit for the $2\%$ to $10\%$ dataset, and $1\%$ degree achieves the best performance on the full dataset. QNLI also has similar observations. Figure \ref{fig:portion_vs_perf} visualizes the KD performance on MNLI with $1\%$ KD data (i.e., the first row of Table \ref{tab:exp1}). The performance curves form an inverted "U" pattern. Smaller degrees (blue) and extremely large degrees (red) are not the best options for KD. Another conclusion is that our findings in Table \ref{tab:exp1} are not because of the coincident variance since the performance gain is consistent and stable in Figure \ref{fig:portion_vs_perf}.

To show our claim is general, we also apply our methods to other datasets. Results are shown in Table. \ref{tab:exp1.2}. Similar to previous findings, we can conclude that KD prefers larger degrees with smaller datasets. Smaller datasets, such as CoLA (8.5k) and MRPC (3.7k), prefer $50\%$ to $70\%$ degrees. Large datasets, such as QQP (363.8k), prefer the $10\%$ degree, which is the same as MNLI (392.7k). Interestingly, the RTE dataset is also small-scale (2.5k) but prefers the $30\%$ degree, which is smaller than the CoLA (8.5k). This may be because of the task difference. Summarizing all findings, \textcolor{ggreen}{\textbf{we empirically suggest the $30\%$ degree for datasets with a moderate size (10k to 100k), $50\%$ degree for datasets with a small size (less than 10k) and $10\%$ degree for datasets with a large size (more than 100k)}}.

\begin{minipage}[t]{0.48\textwidth}
    % \begin{figure}
        \centering
        \includegraphics[width=\textwidth]{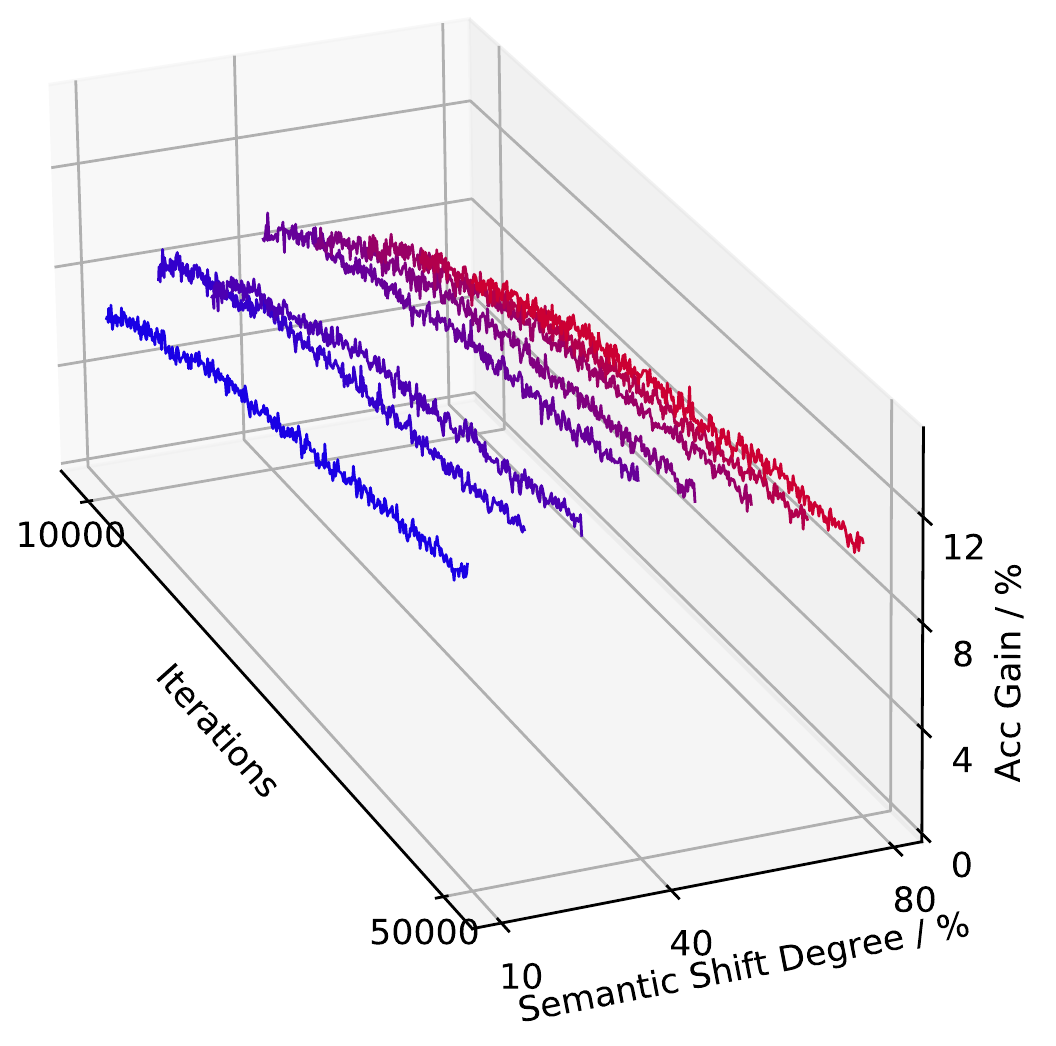}
        \captionof{figure}{KD performances with different proportions of changed tokens. $1\%$ MNLI training set serves as KD data, and the MNLI-M validation set is used for evaluation.}
        \label{fig:portion_vs_perf}
    % \end{figure}
\end{minipage}
\hfill
\begin{minipage}[t]{0.48\textwidth}
    \centering
        \includegraphics[width=\textwidth]{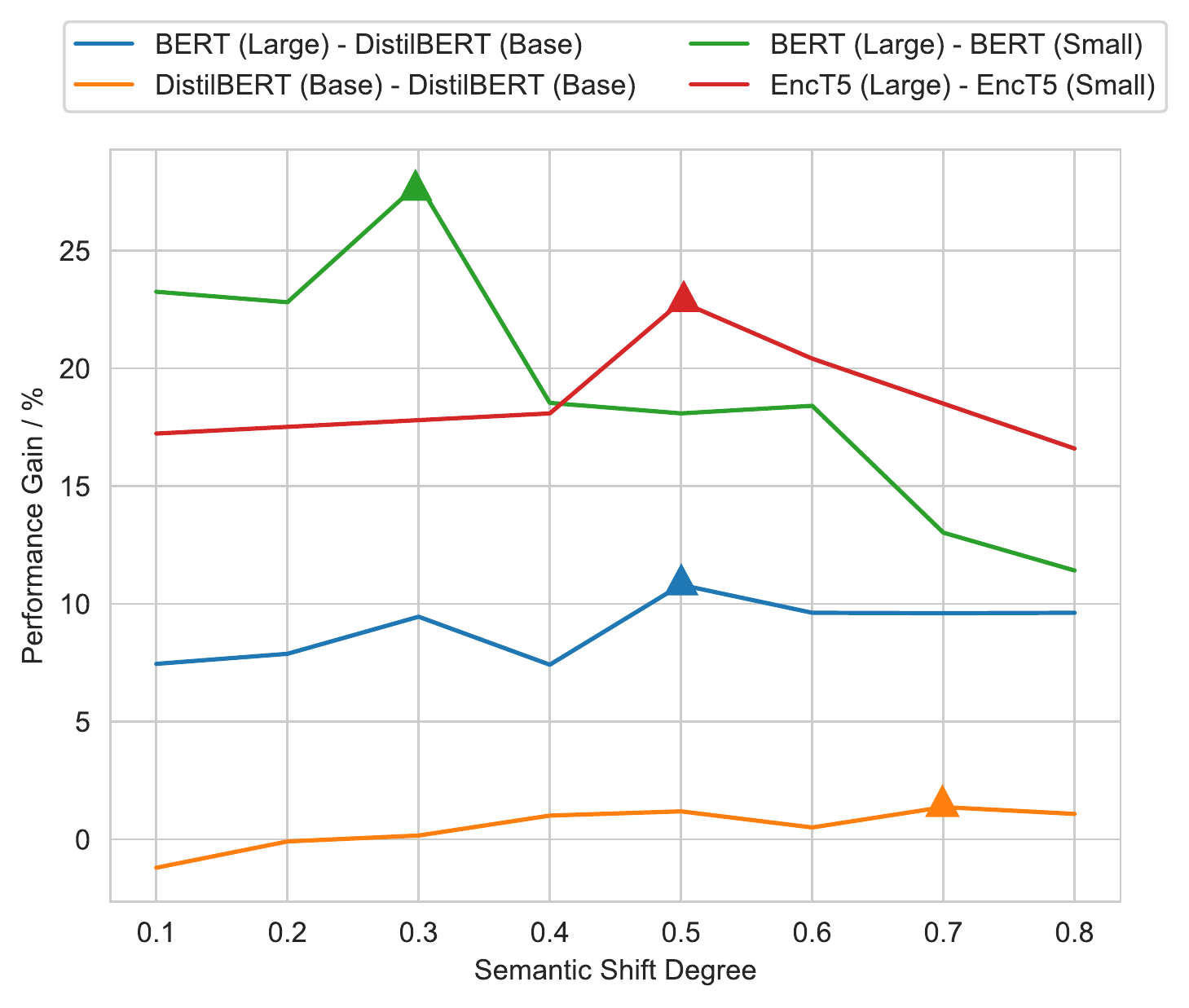}
        \captionof{figure}{KD performance with different models and semantic shift degrees on the CoLA dataset. Triangle markers denote the peak of each curve.}
        \label{fig:self-distill}
\end{minipage}

\subsection{Analysis}
\label{sec:analysis}
\textbf{Different teacher-student models} To show our findings also hold for other teacher-student model pairs. We conduct experiments with three more different teacher-student model pairs: (1) BERT (Large) - BERT (Small): We keep the teacher model unchanged (336M) but use a smaller student model (29.1M) \citep{turc2019well}. (2) DistilBERT (Base) - DistilBERT (Base): We keep the student model unchanged but use a smaller teacher model (itself), which is also called self-distillation. (3) EncT5 (Large) - EncT5 (Small): We change the model architecture from BERT to the T5 family. EncT5 \cite{liu2021enct5} is a variant of T5 \cite{raffel2020exploring} that removes all decoder layers of T5 and only uses T5 encoders and a pooling layer to do text classification tasks. EncT5 has almost the same performance as T5 but is more computationally friendly (half of the parameters); therefore, we use EncT5 in our experiments. EncT5 (Large) has 354M parameters, and EncT5 (Small) only has 37M parameters\footnote{EncT5 experiments use 8 TPU v3 slices since 1x V100 16G does not have enough GPU memories.}. Following previous experiment settings, we use kNN ($k=15$) to augment data and the CoLA dataset for the experiment. Figure \ref{fig:self-distill} shows their KD performance gain (compared with vanilla KD) with different semantic shift degrees. We conclude from the figure that all teacher-student models prefer larger semantic shift degrees (more than $10\%$, which is the best option for fine-tuning). However, different models may prefer different degrees. For example, the $30\%$ degree will reach the highest performance on BERT (Large) - BERT (Small),  and the $50\%$ degree performs best on EncT5 (Large) - EncT5 (Small). Besides, extremely large degrees such as $80\%$ will cause a performance drop in all models. The findings are the same as Section \ref{sec:kdprefer}.

\textbf{Different data augmentation methods.} Our conclusions are not limited to the kNN. We use two more data augmentation methods to support this claim: (1) Easy Data Augmentation (EDA) \cite{wei-zou-2019-eda}: EDA is a combination of four lexical-level augmentation methods, i.e., synonym replacement, random insertion, random swap, and random deletion. (2) Random Replacement: Replace tokens with other random tokens in the vocabulary. We use $1\%$ MNLI training data for distillation, the MNLI-M validation set for evaluation, accuracy as the metric, and report results in Table \ref{tab:diffda}. We can see that EDA prefers $30\% \sim 40\%$ degrees, which is similar to the kNN. Random replacement, however, prefers the $20\%$ degree. Random replacement replaces tokens with random tokens, whereas kNN replaces tokens with similar tokens. Therefore, the random replacement has more semantic shifts than the kNN, even if the proportion of changed tokens is the same, making the best degree shifts to a smaller one. Besides, extremely large degrees harm performances, as previously observed.

\begin{table}
    \centering
    
    \caption{Results of KD performance with data augmentation methods and semantic shift degrees.}
    \vspace{1mm}
    \resizebox{\textwidth}{!}{
    \begin{tabular}{c|ccccccccc}
    \toprule
    \multirow{2}{*}{\textbf{Methods}} &\multicolumn{9}{|c}{\textbf{Semantic Shift Degree}} \\
     & \textbf{0\% (Vanilla KD)} & \textbf{10\%} & \textbf{20\%} & \textbf{30\%} & \textbf{40\%} & \textbf{50\%} & \textbf{60\%} & \textbf{70\%} & \textbf{80\%} \\
     \midrule
      \multicolumn{10}{c}{\textbf{MNLI (392.7k) (Evaluated on the MNLI-M validation set)}} \\
    \midrule
     \textbf{kNN} & \multirow{3}{*}{$68.89$} & $76.39$	& $77.31$& 	$77.17$	& $\textcolor{ggreen}{\bm{78.06}}$	&$77.47$	&$77.02$&	$76.23$&	$75.31$\\
     \textbf{EDA} & & $75.81$	&$	76.83$	&$	\textcolor{ggreen}{\bm{77.48}}$	&$	77.22$	&$	76.88$	&$	76.88$	&$	76.15$	&$	75.8$\\
     \textbf{Random Replacement} & & $75.92$	&$\textcolor{ggreen}{\bm{76.94}}$	&$76.86$	&$75.98$	&$74.91$	&$73.43$	&$	72.27$	&$	71.6$\\
    \bottomrule
    \end{tabular}
    }
    \label{tab:diffda}
\end{table}

\begin{table}
    \centering
    \caption{Showcase of the augmented data (kNN) and synthesized data (back translation). Examples are selected from the MNLI dataset. Highlight tokens are changed tokens compared with the original input. kNN with a larger semantic shift degree changes more tokens and semantics.}
    \vspace{1mm}
    \resizebox{\textwidth}{!}{
    \small
    \begin{tabular}{c|c}
    \toprule
      \multicolumn{2}{c}{\textbf{Original Input}} \\
    \midrule
     \multicolumn{2}{c}{\texttt{\textbf{premise:} How do you know? All this is their information again.}} \\
     \multicolumn{2}{c}{\texttt{\textbf{hypothesis:} This information belongs to them.}} \\
     \midrule
     \multicolumn{1}{c|}{\textbf{kNN (k=15). 10\% Semantic Shift Degree.}} & \multicolumn{1}{c}{\textbf{kNN (k=15). 30\% Semantic Shift Degree.}} \\
    \midrule
    \multicolumn{1}{l|}{\texttt{\textbf{premise:} \makecell[l]{\textcolor{red}{where} do you know? all this is \\ \textcolor{red}{our} information again.}}} & \multicolumn{1}{l}{\texttt{\textbf{premise:} \makecell[l]{\textcolor{red}{where did} you know? \textcolor{red}{and -} is \\ their \textcolor{red}{knowledge} again \textcolor{red}{of}}}}\\
     \multicolumn{1}{l|}{\texttt{\textbf{hypothesis:} \textcolor{red}{all} information belongs to \textcolor{red}{it}.}} &  \multicolumn{1}{l}{\texttt{\textbf{hypothesis:} this information belongs to \textcolor{red}{her}.}}\\
     \midrule
     \multicolumn{1}{c|}{\textbf{kNN (k=15). 70\% Semantic Shift Degree.}} & \multicolumn{1}{c}{\textbf{Synthesized Data (Back Translation)}} \\
    \midrule
      \multicolumn{1}{l|}{\texttt{\textbf{premise:} \makecell[l]{\textcolor{red}{590 go themango}? all \textcolor{red}{these} \\ is \textcolor{red}{a 406 still -}}}} & \multicolumn{1}{l}{\texttt{\textbf{premise:} You got it?}}\\
     \multicolumn{1}{l|}{\texttt{\textbf{hypothesis:} \textcolor{red}{which 296} belongs \textcolor{red}{with him}.}} & \multicolumn{1}{l}{\texttt{\textbf{hypothesis:} This is their information.}}\\
      \bottomrule
    \end{tabular}
    }
    \label{tab:case}
\end{table}

\textbf{Showcase of augmented data.} Table \ref{tab:case} shows the augmented data and synthesized data generated by kNN and back translation. We can see that the $30\%$ semantic shift degree destroys the semantics to some extent but still retains the main semantics. The $70\%$ degree, however, will make the input meaningless, causing the OOD problem and performance drop. Synthesized data has a much better quality. However, they are too expensive to be produced.
\section{Applications}
\label{sec:app}

\begin{wraptable}{r}{0.3\textwidth}
    \centering
    \caption{Statistical data of some widely used biomedical datasets.}
    \vspace{1mm}
    \resizebox{0.3\textwidth}{!}{
    \begin{tabular}{cccc}
    \toprule
         \textbf{Datasets} & \textbf{Training} & \textbf{Test} \\
    \midrule
         AIMed \cite{bunescu2005comparative} & 4938 & 549\\
         BioInfer \cite{pyysalo2007bioinfer} & 8544 & 950\\
         LLL \cite{hakenberg2005lll}& 300 & 34\\
         DDI \cite{herrero2013ddi}& 2937 & 979\\
         ChemProt \cite{krallinger2017overview} & 4154 & 3458\\
         GAD \cite{bravo2015extraction}& 4796 & 534\\
    \bottomrule
    \end{tabular}
    }
    \label{tab:biodata}
\end{wraptable}

One application of our findings is to better distill biomedical knowledge from large language models. Biomedical datasets usually require human experts to annotate, making them much smaller than common sense datasets. Table \ref{tab:biodata} shows some widely used biomedical datasets that aim to extract protein-protein interactions, drug-drug interactions, disease-gene relationships, or chemical-protein relations. Most of them only contain thousands or hundreds of data. Therefore, distilling biomedical knowledge from large-scale models to small-scale models is challenging. To show that our findings help distill biomedical knowledge, we choose GAD \cite{bravo2015extraction} to conduct experiments. GAD requires models to identify whether one disease has a relation to a specific gene from a given text. To avoid models inferring the answer from the name of the disease and gene rather than the given text, target entities have been masked with "@GENE\$" and "@DISEASE\$". For example, the given text "this study proposes that A/A genotype at position -607 in @GENE\$ gene can be used as a new genetic maker in Thai population for predicting @DISEASE\$ development." has a label $1$ (indicating there is a relation between the disease and gene). We use the same training/test splits as previous works \cite{lee2020biobert, sarrouti2022comparing}. We select BERT (Large) as the teacher model and BERT (Small) as the student model. kNN ($k=15$) is used to augment data. Table \ref{tab:biores} shows the KD results. We can conclude that augmented data benefits KD, and KD with larger degrees ($20\% \sim 40\%$) enables the student model to perform similarly or better than the teacher model ($80.67\%$ accuracy), showing the effectiveness of our findings. Moreover, extremely larger degrees ($70\% \sim 80\%$) will harm the KD and make the student model perform even worse than vanilla fine-tuning ($78.55\%$ accuracy).

\begin{table}[]
    \centering
    \caption{KD performance (accuracy) on the GAD dataset. The teacher model has a $80.67\%$ accuracy after fine-tuning. Fine-tuning the student model can only have $78.55\%$ accuracy. Vanilla KD improves student model's performance, and KD with larger degrees could even make the student model perform similarly ($20\%$ and $40\%$) to the teacher model or even better ($30\%$) than the teacher model.}
    \vspace{1mm}
   \resizebox{\textwidth}{!}{
    \begin{tabular}{c|ccccccccc}
    \toprule
    \multirow{2}{*}{\textbf{Method}} &\multicolumn{9}{|c}{\textbf{Semantic Shift Degree}} \\
     & \textbf{0\% (Vanilla KD)} & \textbf{10\%} & \textbf{20\%} & \textbf{30\%} & \textbf{40\%} & \textbf{50\%} & \textbf{60\%} & \textbf{70\%} & \textbf{80\%} \\
    \midrule
     \textbf{kNN} & $79.59$ & $79.45$& $80.08$ &$\textcolor{ggreen}{\bm{81.11}}$&$80.78$&$79.67$&$79.74$&$79.24$&$78.08$\\
    \bottomrule
    \end{tabular}
    }
    \label{tab:biores}
\end{table}
\section{Related Work}

\textbf{Knowledge Distillation} Knowledge distillation is first proposed by \citep{hinton2015distilling}. KD is often used to distill knowledge from large-scale to small-scale models to boost the small-scale models' performance. The main idea is to let student models learn from teacher models' output label probability distribution by minimizing KL divergence \citep{hinton2015distilling} or mean square errors \citep{liangmixkd}. In principle, KD is model-agnostic and could be applied between arbitrary models. For example, \cite{tang2019distilling} distills knowledge from BERT \citep{devlin-etal-2019-bert} to BiLSTM. Model-specific KD attracts more attention with the rise of transformers \citep{NIPS2017_3f5ee243}. \cite{sun2019patient} conducts KD not only on the output layer but also on the hidden layers. \cite{sun-etal-2020-mobilebert} introduces attention distillation in transformers. \cite{jiao-etal-2020-tinybert} combines attention distillation and hidden layer distillation. In our work, we use model-agnostic KD to make our conclusions more general.

\textbf{Data Augmentation and Synthesized Data} Classic data augmentation methods are usually cheap and focus on the token level. Synonym replacement \citep{kolomiyets-etal-2011-model} replaces tokens with synonyms. K-nearest-neighbors \citep{wang2015s} finds similar tokens from embeddings and then conducts replacement. Easy data augmentation \citep{wei-zou-2019-eda} combines synonym replacement, random swap, random insertion, and random deletion. TreeMix \citep{zhang-etal-2022-treemix} uses a constituency parser to mix two instances and generate a new one. On the contrary, synthesized data is usually generated by autoregressive models \citep{brown2020language, openai2022chatgpt}, making it expensive but more diversified. \cite{zhou-etal-2022-flipda} uses T5 \citep{raffel2020exploring} to modify the original input and generate data that flips the label. \cite{yoo-etal-2021-gpt3mix-leveraging} uses in-context learning to guide GPT-3 \citep{brown2020language} to generate more task-specific data. Synthesized data is usually preferred in the few-shot learning setting due to the high cost \citep{yoo-etal-2021-gpt3mix-leveraging, zhou-etal-2022-flipda}.

\textbf{Data Augmentation in Knolwegde Distillation} Although most augmented data and synthesized data are initially designed for fine-tuning settings, they can directly be applied in KD. However, they may be sub-optimal for KD. \cite{wang2022makes} finds that good KD augmented data should have a low entropy of the teacher model's output. They find dropping augmented data with high entropy could benefit KD. Our main difference is that we focus on adapting a data augmentation paradigm initially designed for fine-tuning settings to KD, whereas they focus on selecting a subset of augmented data generated by a specific data augmentation paradigm that can benefit KD most.
\section{Conclusion and Future Work}
We investigate what data augmentation methods KD prefers: (1) Cheap augmented data with online augmentation to ensure sufficient quantity, and (2) larger semantic shift degrees. Moreover, we show smaller datasets prefer larger semantic shift degrees in KD. These findings give guidance on how to use data augmentation for KD. Our work shows that KD has different preferences for data augmentation compared with fine-tuning, and a proper selection could improve KD results consistently. Therefore, we encourage the community to explore more KD-specific data augmentation methods. In the future, we will extend our findings from text classification to more tasks such as vision tasks, graph tasks, and natural language generation tasks.

\section*{Limitations}
We use the proportion of changed tokens to reflect the semantic shift degree. Though intuitive, it is not comprehensive enough. The discussion of Random Replacement in Section \ref{sec:analysis} could reveal this problem. Therefore, a more comprehensive way to represent semantic shifts is needed. 

\section*{Acknowledgement}
We thank Yangsibo Huang for the help with the experiments. This research is based upon work supported in part by U.S. DARPA KAIROS Program No. FA8750-19-2-1004 and U.S. DARPA AIDA Program No. FA8750-18-2-0014. The views and conclusions contained herein are those of the authors and should not be interpreted as necessarily representing the official policies, either expressed or implied, of DARPA, or the U.S. Government. The U.S. Government is authorized to reproduce and distribute reprints for governmental purposes notwithstanding any copyright annotation therein.

\bibliography{anthology,custom}
\bibliographystyle{plain}

\appendix
\newpage
\section{Proof}
\label{appendix:thm}

\setcounter{theorem}{0}
\begin{theorem}

Assume a data distribution $\mathcal{D}=\{(x, y)\}$ with $x\in \mathbb{R}^d \sim \mathcal{N}(0, I)$ where $d\geq 2$, and $y \sim \mathcal{B}(f(x))$. We also assume a Lipschitz bound of B on $P(x)$. Training data $ D_{train}$ and testing data $\mathcal{D}_{test}$ are $n$ and $m$ i.i.d. sample from $\mathcal{D}$, respectively. We also assume access to a teacher model $g(x)$, so that $\forall x, |g(x)-f(x)|<\epsilon_t$. The teacher model is difference from training distribution in that it does not have information of input density $p(x)$.

We assume for each training sample $(x,y)$ individually, we augment its information to span a local gaussian distribution $\mathcal{N}(x, \tau I)$. We then i.i.d. sample $m$ datum points from the following distribution:

\begin{align*}
    \mathcal{P}_{aug}(x') &= \frac{1}{n}
    \sqrt{\frac{1}{(2\pi)^d\tau}}
    \sum_{x\in D_{train}}
    e^{-\frac{\|x'-x_i\|_2^2}{2\tau}} \\
    y' &=g(x'),
\end{align*}

and get the augmented dataset $\mathcal{D}_{aug}$.

Let $x_u=2d\ln(2\pi d) + 12\ln(2) - 4\ln(\epsilon)$, as long as

\begin{align*}
    \tau <& \frac{2^{d-3}\epsilon}{x_u^{d+2}} \\
    n >& \frac{16}{\epsilon^2} \left((d^2+5d+6)\ln2 + (d^2+d)\ln x_u + d\ln d - d\ln\epsilon - d\ln\delta-\frac{d}{2}\ln\tau \right)
\end{align*}

we have with $1-\delta$ probability, for any $h\in\mathcal{H}$:

\begin{equation}
    \epsilon_T(h) - \epsilon_S(h) < \epsilon + \lambda,
\end{equation}

where $\lambda=\epsilon_T(h^\star)+\epsilon_T(h^\star)$.

\end{theorem}

\paragraph{Proof of Theorem~\ref{thm:teacher_augmentation}}
\begin{proof}

We start from bounding 
\[
    d_{\mathcal{H}\Delta\mathcal{H}}( D_{train}, \mathcal{D}_{aug}) \leq 
    2 \|P_{train}(x) - P_{aug}(x)\|_1.
\]

We set a ball occupying $1-\frac{\epsilon}{4}$ of $P_{train}$ mass, which has radius 
\[
    4\ln \left(\frac{8(2\pi)^\frac{d}{2}}{\epsilon\Gamma(\frac{d}{2})} (\frac{d-2}{e})^{(\frac{d}{2}-1)} \right) 
    \leq 2d\ln(2\pi d) + 12\ln(2) - 4\ln(\epsilon) := x_u
\]

Outside the ball, with $1-\frac{\delta}{2}$ probability, no more than $\frac{\epsilon}{4}+\sqrt{\frac{2}{n}\ln\frac{2}{\delta}}$ proportaion of all training data appears.

$\hat{P}(x)$ has Lipschitz bound of
\[
    B' \leq \frac{n}{\sqrt{\tau}} e^{-\frac{1}{2}} \leq \frac{n}{\sqrt{\tau}}
\]

Each cube of size $r$ has cumulative error from its center of at most $\frac{1}{2}(B+B')Vr$, we require a grid size of at most size 
\[r=\frac{\epsilon}{2(B+B')(2x_u)^d}\]
to control the overall cumulative error below $\frac{\epsilon}{4}$. This grid size results number of grids
\[
    n_g \geq \left(2\frac{x_u}{r}\right)^d = \left(\frac{4(B+B')(2x_u)^{d+1}}{\epsilon}\right)^d
\]

For each grid center $x$, by applying Hoeffding's inequality, with $1-\frac{\delta}{2n_g}$ probability:

\begin{equation}
    \left|\frac{1}{n} \sum_{x'\in  D_{train}} e^{-\frac{\|x-x'\|_2^2}{2\tau}} - \mathbb{E}_{x'\sim \mathcal{D}}  e^{-\frac{\|x-x'\|_2^2}{2\tau}}\right|
    \leq \sqrt{\frac{1}{2n} \ln \frac{2n_g}{\delta}}.
\end{equation}

We now prove that the second term approaches ground truth $P(x)$ when $\tau \rightarrow 0$.

\begin{align*}
    \mathbb{E}_{x'\sim \mathcal{D}}  e^{-\frac{\|x-x'\|_2^2}{2\tau}}
    &= \int_{x'\in\mathbb{R}^d} e^{-\frac{\|x-x'\|_2^2}{2\tau}} e^{-\frac{\|x'\|_2^2}{2}} dx' \\
    &= \int_{x'\in\mathbb{R}^d} e^{-\frac{1}{2}((1+\frac{1}{\tau})x'^\top x' -2 \frac{1}{\tau} x'^\top x  + \frac{1}{\tau} x^\top x)} dx' \\
    &= \int_{x'\in\mathbb{R}^d} e^{-\frac{1}{2}(\frac{1+\tau}{\tau}\|x' - \frac{1}{1+\tau}x\|_2^2 + \frac{1}{1+\tau} x^\top x)} dx' \\
    &= \sqrt{\frac{\tau(2\pi)^d}{1+\tau}} e^{-\frac{1}{2(1+\tau)} \|x\|_2^2}.
\end{align*}

Therefore, 
\begin{align*}
    \left\|\sqrt{\frac{1}{(2\pi)^d\tau}}\mathbb{E}_{x'\sim \mathcal{D}}  e^{-\frac{\|x-x'\|_2^2}{2\tau}} - e^{-\frac{1}{2(1+\tau)} \|x\|_2^2} \right\|_\infty \leq \frac{\tau(x_u^2+1)}{2}
\end{align*}

When 
\begin{align*}
    \tau <& \frac{2^{d-3}\epsilon}{x_u^{d+2}} \\
    n >& \frac{16}{\epsilon^2} \left((d^2+5d+6)\ln2 + (d^2+d)\ln x_u + d\ln d - d\ln\epsilon - d\ln\delta-\frac{d}{2}\ln\tau \right)
\end{align*}

we have
\[
    \|\hat{P}-P\|_1 < \epsilon
\]

and 

\[
    (\epsilon_T - \epsilon_s)(h) < \epsilon+\lambda
\]

\end{proof}
%%%%%%%%%%%%%%%%%%%%%%%%%%%%%%%%%%%%%%%%%%%%%%%%%%%%%%%%%%%%

\end{document}